\documentclass[journal]{IEEEtran}

   \usepackage{graphicx}
   \usepackage{amsfonts}
   \usepackage{amsmath}
   \usepackage{multirow}
   \usepackage{color}

\begin{document}

\title{RelationTrack: Relation-aware Multiple Object Tracking with Decoupled Representation}

\author{En Yu, Zhuoling Li, Shoudong Han and Hongwei Wang

\thanks{E. Yu and Z. Li contribute equally to this work (Corresponding author: Shoudong Han).}
\thanks{E. Yu, S. Han and H. Wang are with the National Key Laboratory of Science and Technology on Multispectral Information Processing, School of Artificial Intelligence and Automation, Huazhong Univerisity of Science and Technology, 1037 Luoyu Road, Wuhan, China, PC 430074 (e-mail:\{yuen, shoudonghan, hongweiwang\}@hust.edu.cn)}
\thanks{Z. Li is with the Shenzhen International Graduate School, Tsinghua University, Ministry of Education, Shenzhen, China, PC 518000 (e-mail: lzl20@mails.tsinghua.edu.cn).}}

\maketitle

\begin{abstract}
Existing online multiple object tracking (MOT) algorithms often consist of two subtasks, detection and re-identification (ReID). In order to enhance the inference speed and reduce the complexity, current methods commonly integrate these double subtasks into a unified framework. Nevertheless, detection and ReID demand diverse features. This issue would result in an optimization contradiction during the training procedure. With the target of alleviating this contradiction, we devise a module named Global Context Disentangling (GCD) that decouples the learned representation into detection-specific and ReID-specific embeddings. As such, this module provides an implicit manner to balance the different requirements of these two subtasks. Moreover, we observe that preceding MOT methods typically leverage local information to associate the detected targets and neglect to consider the global semantic relation. To resolve this restriction, we develop a module, referred to as Guided Transformer Encoder (GTE), by combining the powerful reasoning ability of Transformer encoder and deformable attention. Unlike previous works, GTE avoids analyzing all the pixels and only attends to capture the relation between query nodes and a few self-adaptively selected key samples. Therefore, it is computationally efficient. Extensive experiments have been conducted on the MOT16, MOT17 and MOT20 benchmarks to demonstrate the superiority of the proposed MOT framework, namely RelationTrack. The experimental results indicate that RelationTrack has surpassed preceding methods significantly and established a new state-of-the-art performance, e.g., IDF1 of 70.5\% and MOTA of 67.2\% on MOT20.
\end{abstract}


\section{Introduction}

\IEEEPARstart{A}{s} a fundamental vision task, multiple object tracking (MOT) aims to estimate the locations of several targets \cite{liang2019small, kang2017t} and identify which of them belong to the same object \cite{sun2020survey, zhang2020person, ciaparrone2020deep, weng2020gnn3dmot, xiang2015learning}. Much attention has been drawn due to its numerous practical applications, such as video analysis \cite{takahashi2017aenet}, autonomous driving \cite{luo2018fast}, robots \cite{manglik2019forecasting}, etc. Although prominent progress has been achieved, existing MOT systems still suffer from poor tracking precision and need improvements.

\begin{figure}[tb]
    \centering
    \includegraphics[scale=0.42]{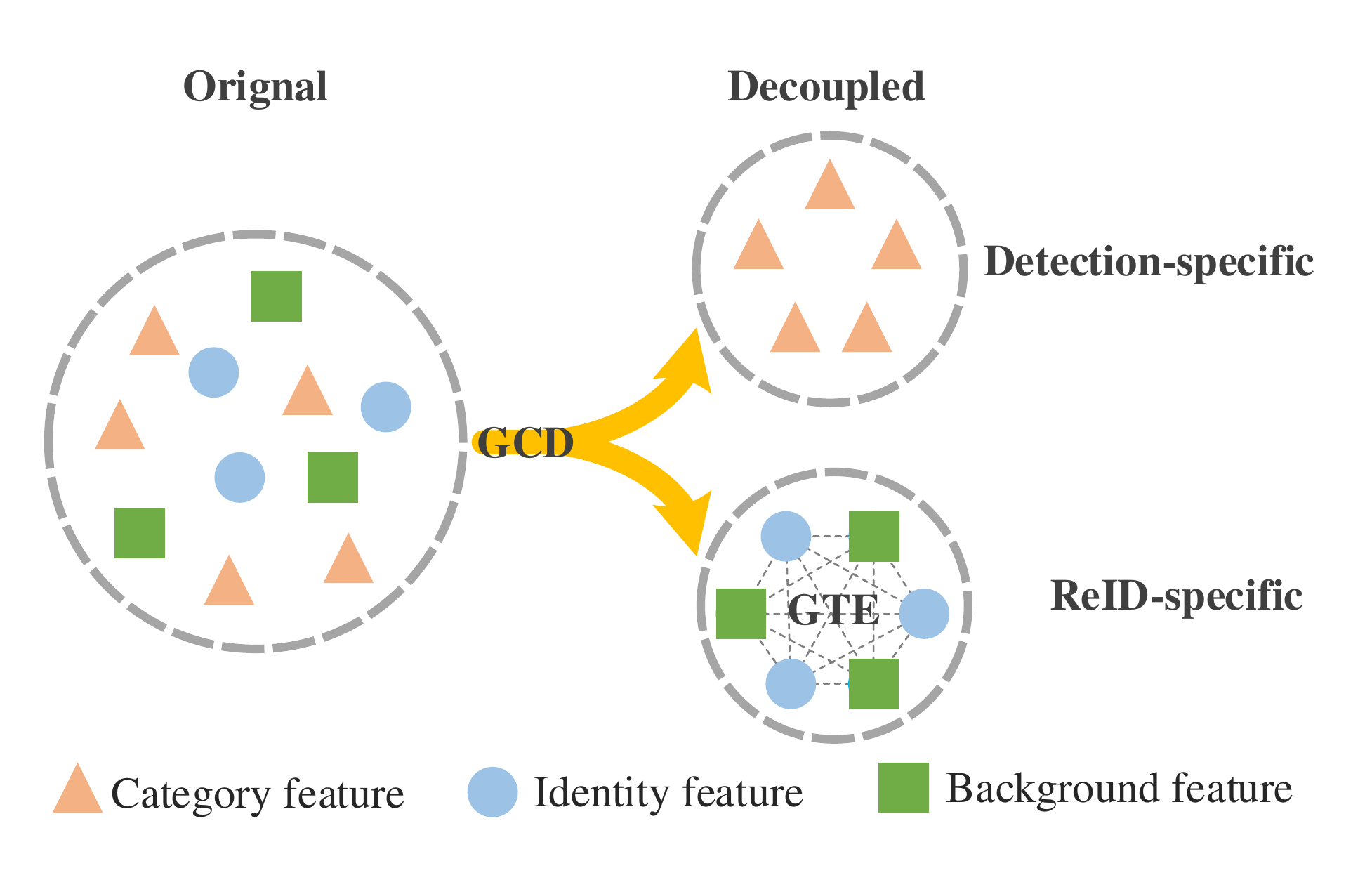}
    \caption{Diagram that presents how the proposed two modules (GCD and GTE) affect the training procedure. GCD can decouple the learned features as detection-specific and ReID-specific embeddings. Instead of using dot-product attention of transformer that would lead to huge computational cost, GTE combine deformable attention and transformer encoder to capture the global semantic relation.}
    \vspace{-0.4cm}
\end{figure}

Former MOT frameworks mainly comprise two sub-models, a detection model to localize the targets and a re-identification (ReID) model for connecting them to the trajectories \cite{zhan2020simple}. However, executing these two models separately results in slow inference speed and huge computational cost. A possible solution to this problem is building networks as the joint-detection-and-embedding architecture \cite{wang2019towards}, which incorporates detection and ReID into a single network and conducts them simultaneously.

Nevertheless, directly merging the detection and ReID models into a single framework leads to a serious optimization contradiction \cite{liang2020rethinking}. For the detection part, the network wishes to strengthen the representation similarity of objects belonging to the same category. By contrary, the ReID part desires to maximize the feature discrepancy among various targets, even though they pertain to an identical category. Their inconsistent optimization objectives hinder current MOT frameworks evolving towards more efficient forms.

In order to address this contradiction, we design a self-motivated feature decoupling module, Global Context Disentangling (GCD) that decouples the feature representation as detection-specific and ReID-specific embeddings, as shown in Fig. 1. Verified by our experiments, this module contributes to alleviating the contradiction between detection and ReID, and it improves the tracking precision significantly (e.g., from 73.3\% to 74.9\% on MOT17 for the metric IDF1 as illustrated in Table IV).

Additionally, we observe that previous methods often track targets with only local information. However, a prior sense behind MOT is that the global relation among objects and background is important since the surrounding pixels are efficient cues for tracking  \cite{zhang2020relation}. To capture this long-range relation, a possible solution is employing the global attention \cite{yin2020disentangled}. Nevertheless, global attention needs to compute the pairwise similarity of every query nodes with all the other pixels in the image to generate an attention map. This strategy brings a severe calculation burden.

We argue that not all the pixels affect the semantic content of query nodes. Therefore, only considering the relation with a small handful of crucial key samples could be a better alternative. With respect to this hypothesis, we employ deformable attention \cite{zhu2020deformable} to incorporate the structural relation. Compared with global attention, deformable attention is quite lightweight and reduces the computational complexity from $O(n^{2})$ to $O(n)$. Besides, unlike graph based methods that only gather information from restricted surrounding pixels \cite{weng2020gnn3dmot}, deformable attention selects valuable key samples automatically across the whole image.

Furthermore, we resort to the powerful reasoning ability of Transformer encoder \cite{carion2020end, vaswani2017attention} for better modeling the long-range dependency. Through combining the deformable attention and Transformer encoder, the resulted module, Guided Transformer Encoder (GTE), allows the MOT framework (RelationTrack) to explore the rich content of pixel-to-pixel relation with a global receptive field.

To demonstrate the superiority of RelationTrack\footnote{Code will be available online once the paper is accepted.}, extensive experiments have been conducted on three benchmark datasets, i.e., MOT16 \cite{milan2016mot16}, MOT17 \cite{milan2016mot16} and MOT20 \cite{dendorfer2020mot20}. The results indicate that the proposed framework has outperformed preceding counterparts significantly. For instance, with respect to the metric IDF1, RelationTrack has surpassed the former state-of-the-art (SOTA) method FairMOT \cite{zhan2020simple} by 3.0\% on MOT16 and 2.4\% on MOT17.

Comprehensively, our contributions are summarized as follows:

\vspace{1mm}
$\bullet$$\ $We observe that the optimization contradiction between detection and ReID during the training procedure would hinder networks evolving towards more efficient forms. To address this contradiction, we devise a self-motivated module named GCD that decouples the learned features as detection-specific and ReID-specific embeddings.

\vspace{1mm}
$\bullet$$\ $We highlight the importance of global relation among objects and background for ReID of MOT. By combining the advantages of deformable attention and Transformer encoder, we develop a lightweight module (GTE) for exploring the long-range dependency across the whole image.

\vspace{1mm}
$\bullet$$\ $ Incorporating the power of GCD and GTE, the proposed MOT framework, RelationTrack, surpasses its previous counterparts obviously. Evaluated with 5 groups of experiments on 3 benchmark datasets,  RelationTrack establishes a new SOTA performance. For example, \textbf{we achieve \textbf{IDF1} of \textbf{70.5\%} and \textbf{MOTA} of \textbf{67.2\%} on the MOT20 benchmark.}

\section{Related Works}
Influenced by recent great progresses of detection techniques \cite{cai2018cascade, law2018cornernet, yang2019reppoints}, detection-based MOT algorithms have dominated the mainstream. These algorithms mostly
comprise two parts, i.e., estimating the locations of targets and associating them to the trajectories. According to how the framework is organized, existing detetion-based MOT methods can mainly be categorized into three classes, which are introduced as follows.

\vspace{-2mm}
\subsection{Tracking-by-detection}
There exist numerous publications following the tracking-by-detection paradigm \cite{zhan2020simple, wang2019towards, chen2018real, yu2016poi, zhu2018online, bewley2016simple, wojke2017simple}. Many of them concentrate on how to enhance the association ability of methods. Early works often address this challenge through designing algorithms based on kinematics, such as Kalman filtering \cite{welch1995introduction}. These methods usually first take current states of concerned targets as input and attempt to predict their locations in the next frames. Afterwards, the Hungarian algorithm \cite{kuhn1955hungarian} would be applied to adjust the predicted results.

Nevertheless, since the trajectories of moving objects (such as pedestrians) are highly diverse and hard to be predicted, kinematics based methods often fail to track the targets. To overcome this obstacle, appearance based strategies are introduced. For example, DeepSort \cite{wojke2017simple} utilizes techniques of ReID to extract features and compute the similarity. FGAGT \cite{shan2020fgagt} considers both motion and appearance information through the graph neural network.

Although with competitive performance, the aforementioned models still suffer from some restrictions. For instance, they usually implement detection and association as two independent sub-models. Both them will influence the final tracking precision considerably. Therefore, if any sub-model fails to behave well, the final results could be terrible. In addition, since the two sub-models do not share network layers, the resulted slow inference speed and heavy computation burden restrict the tracking-by-detection paradigm from further improvements.

\begin{figure*}[htbp]
    \centering
    \includegraphics[scale=0.51]{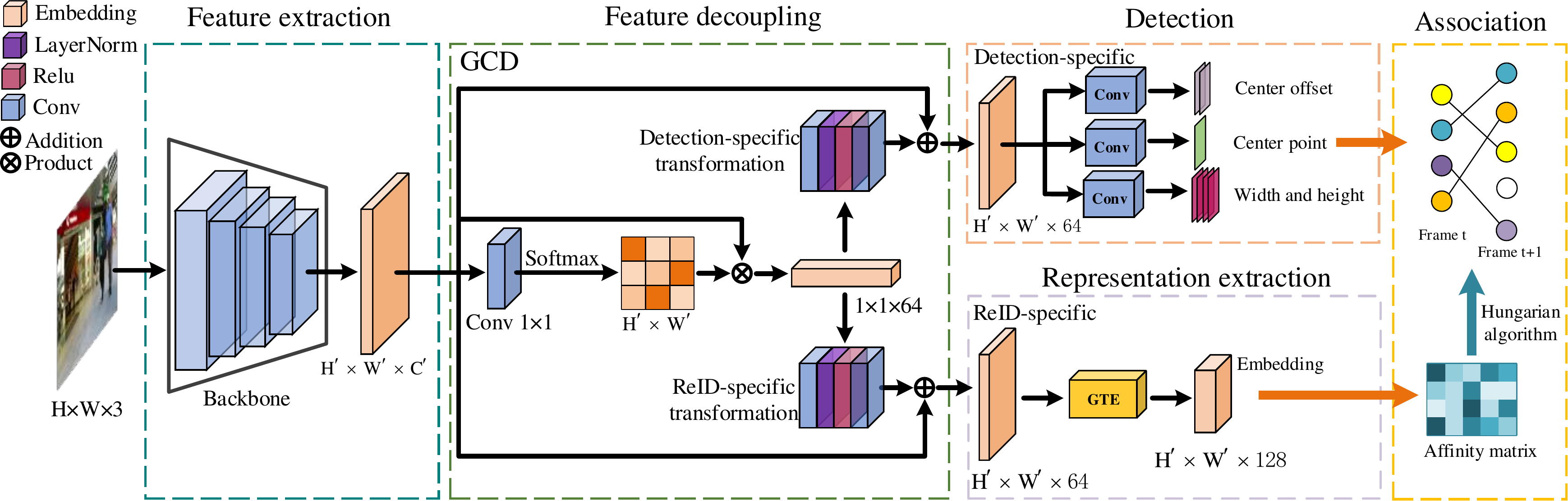}
    \caption{The overall pipeline of RelationTrack (LayerNorm: layer normalization, Relu: rectified linear unit, Conv: convolution).}
    \vspace{-0.4cm}
\end{figure*}

\vspace{-2mm}
\subsection{Joint-detection-and-prediction}
Many attempts have been conducted to overcome the disadvantages brought by implementing two separate sub-models \cite{kang2017t, zhu2017flow, zhang2018integrated, feichtenhofer2017detect, bergmann2019tracking}. Among these attempts, incorporating the two parts, i.e., detecting targets and predicting trajectories, into a unified framework is a common practice.

With respect to this motivation, joint-detection-and-prediction methods employ a single network to localize and predict the position of targets and then associate them. For example, Tracktor \cite{bergmann2019tracking}  adopts the bounding box regression module in Faster R-CNN \cite{ren2016faster} to correct the predicted results. CTracker \cite{peng2020chained} exploits rich content among adjacent frames to enhance the regression precision. MAT \cite{han2020mat} regards the information of well-established kinematic models as extra cues to facilitate the estimation procedure.

Moreover, some works take advantage of Siamese network \cite{bertinetto2016fully, tao2016siamese, li2018high, li2019siamrpn++} to explore the feature similarity and estimate trajectories, such as DeepMOT \cite{xu2019deepmot}, which decomposes MOT as several single object tracking (SOT) tasks. Likewise, Centertrack \cite{zhou2020tracking} produces bounding box offsets with the inspiration from CenterNet \cite{zhou2019objects}.

Generally, joint-detection-and-prediction models behave better than the tracking-by-detection paradigm mainly due to their trajectory prediction blocks \cite{zhan2020simple, wang2019towards, chen2018real, yu2016poi, zhu2018online, bewley2016simple, wojke2017simple}. However, when they are applied to sophiscated application scenarios, further improvements are still demanded.

\vspace{-2mm}
\subsection{Joint-detection-and-embedding}
Similar to the above joint-detection-and-prediction strategies, joint-detection-and-embedding methods often implement their two components, detection and identification, as a one-stage network. However, rather than directly estimating the moving offsets, they associate the concerned targets to trajectories based on extracted embeddings. Among these methods, the outstanding ones include JDE (the first real-time MOT system) \cite{wang2019towards}, FairMOT (an anchor-free tracker) \cite{zhan2020simple}, etc.

As the double branches (detection and identification) of joint-detection-and-embedding models contribute to the performance of each other, the tracking precision of trained networks is often competitive. Nevertheless, we observe that there still exist some obstacles which restrict this paradigm. First of all, the contradiction between detection and identification hurts the optimization procedure. Meanwhile, previous frameworks primarily only utilize the local information and ignore the global semantic relation among targets and background regions. In this paper, we propose several strategies to address these obstacles.

\section{Method}
This section explains how RelationTrack is organized. First of all, Subsection A presents the problem formulation. Then, Subsection B describes the overall framework of RelationTrack. Afterwards, Subsection C, D and E introduce the implementation details of modules (GCD, GTE, detection and association) that compose RelationTrack, respectively. Finally, Subsection F provides the detailed optimization objective settings during the training phase.

\vspace{-2mm}
\subsection{Problem Formulation}
RelationTrack aims to detect the concerned objects (detection) and associate the ones with the same identity among various frames to form trajectories (ReID). It consists of three parts, i.e., a detector $\phi(\cdot)$ to localize the targets, a feature extractor $\psi(\cdot)$ for obtaining representative embeddings and an associator $\varphi(\cdot)$ to produce trajectories.

Formally, given an input image $I_{t} \in \mathbb{R}^{H \times W \times C}$, we denote $\phi(I_{t})$ and $\psi(I_{t})$ as $b_{t}$ and $e_{t}$, where $b_{t} \in \mathbb{R}^{k \times 4}$ and $e_{t} \in \mathbb{R}^{k \times D}$. In these definitions, $H$, $W$ and $C$ represent the height, width and number of channels in the input image $I_{t}$, respectively. $k$, $t$ and $D$ are the number of detected targets, index of $I_{t}$ and dimension of embedding vectors. $b_{t}$ and $e_{t}$ severally refer to the coordinates of bounding boxes and corresponding embedding vectors. After detecting the targets and extracting the corresponding embedding vectors, $\varphi(\cdot)$ will link $b_{t}$ in various frames based on $e_{t}$ to generate the final trajectories.

Generally, in order to estimate the trajectories correctly, the following optimization objectives should be satisfied.

$\bullet$ The bounding boxes corresponding to $b_{t}$ should contain accurate targets properly.

$\bullet$ $e_{t}$ should represent the identity information of targets appropriately. Specifically, the extracted embeddings of targets in various frames with the same identity should be more alike to each other in contrasted to the ones belonging to other identities.

Technically, when the two objectives are both fulfilled, the tracking results would be promising even with a simple $\varphi(\cdot)$, such as the Hungarian algorithm.

\vspace{-2mm}
\subsection{Overall Framework}
As illustrated in Fig. 2, RelationTrack is composed of 5 parts, i.e., feature extraction, feature decoupling, detection, representation extraction and association. In the first part, given a video with $N$ frames $I_{t}$ ($t=1,2,...N$), the backbone (DLA-34 \cite{yin2020center}) transforms every frame to its corresponding feature maps, respectively. Then, in the feature decoupling part (GCD), the learned features are decomposed as detection-specific and ReID-specific information to address the aforementioned feature contradiction problem. Afterwards, the networks of the detection branch (similar to Centernet \cite{zhou2019objects}) localize the concerned objects based on the detection-specific information. Meanwhile, GTE in the representation estimation part would encode the ReID-specific information as discriminative representation. With respect to the bounding boxes and obtained representation, we link the detected targets to the final trajectories using Hungarian algorithm in the association part.

\vspace{-2mm}
\subsection{Global Context Disentangling (GCD)}
In this section, we introduce the details of GCD that decouples the features extracted by the backbone as detection-specific and ReID-specific representation. GCD exactly comprises two phases, i.e., producing the global context vector and utilizing this vector to decompose input feature maps.

Denote $x=\{x_{i}\}_{i=1}^{N_{p}}$ as the input feature maps, where $N_{p}=H' \times W'$ ($H'$ and $W'$ are the height and width of input feature maps, respectively). Then, the process of calculating the global context vector $z$ (the first phase) can be expressed as
\begin{align}
z = \sum\limits_{j=1}^{N_{p}} \frac{exp(W_{k}x_{j})}{\sum\limits^{N_{p}}_{m=1}exp(W_{k}x_{m})}x_{j}
\end{align}
where $W_{k}$ represents a learnable linear projection and it is modeled as a $1\times1$ convolution layer in our implementation.

Afterwards, in the second phase, we devise two transformations that decouple $z$ into two task-specific vectors. By adding them to $x$ through broadcast element-wise addition, we obtain the detection-specific embeddings $d=\{d_{i}\}_{i=1}^{N_{p}}$  and ReID-specific embeddings $r=\{r_{i}\}_{i=1}^{N_{p}}$, respectively. This procedure is formulated as follows:
\begin{align}
& d_{i} = x_{i} + W_{d2}ReLU(\Psi_{ln}(W_{d1}z)) \\
& r_{i} = x_{i} + W_{r2}ReLU(\Psi_{ln}(W_{r1}z))
\end{align}
where $W_{d1}$, $W_{d2}$, $W_{e1}$ and $W_{e2}$ denote four learnable matrices. $ReLU(\cdot)$ and $\Psi_{ln}(\cdot)$ represent the rectified linear unit and layer normalization operator \cite{li2021enabling}, respectively. Given a data batch $I$ with the shape of $(B', H', W', C')$, $\Psi_{ln}(\cdot)$ can be defined as
\begin{align}
& \mu_{b} = \frac{1}{H'W'C'} \sum\limits^{H'}_{1} \sum\limits^{W'}_{1} \sum\limits^{C'}_{1} I_{bhwc} \\
& \sigma^{2}_{b} = \frac{1}{H'W'C'} \sum\limits^{H'}_{1} \sum\limits^{W'}_{1} \sum\limits^{C'}_{1} (I_{bhwc} - \mu_{b})^{2} \\
& \tilde{I}_{bhwc} = \frac{I_{bhwc} - \mu_{b}}{\sqrt{\sigma^{2}_{b} + \epsilon}}
\end{align}
where $I_{bhwc}$ and $\tilde{I}_{bhwc}$ are the elements in input and output data batches with the index of $(b, h, w, c)$, and $\epsilon$ denotes a tiny predefined value.

It could be observed from Equation (1) that $z$ is invariant to the selection of $i$ while aggregating the global context information. All the elements of $d$ and $r$ can be computed using the same $z$. Caused by this characteristic, the calculation complexity of GCD is only $O(C^{2})$. In contrast to former global attention methods with complexity of $O(HWC^{2})$ \cite{zhang2020relation}, GCD is quite computationally efficient. Moreover, according to the experiments in Subsection D of Section IV, GCD successfully decouples learned features and addresses the feature contradiction problem.

\vspace{-2mm}
\subsection{Guided Transformer Encoder (GTE)}
Attention is a widely adopted strategy to enhance the discriminability of learned features \cite{zhang2020relation}. However, most previous works generate attention maps by convolutions with limited receptive fields \cite{chen2019mixed}, which ignore the global relation information among various targets and background regions.

\begin{figure}[tbp]
    \centering
    \includegraphics[scale=0.55]{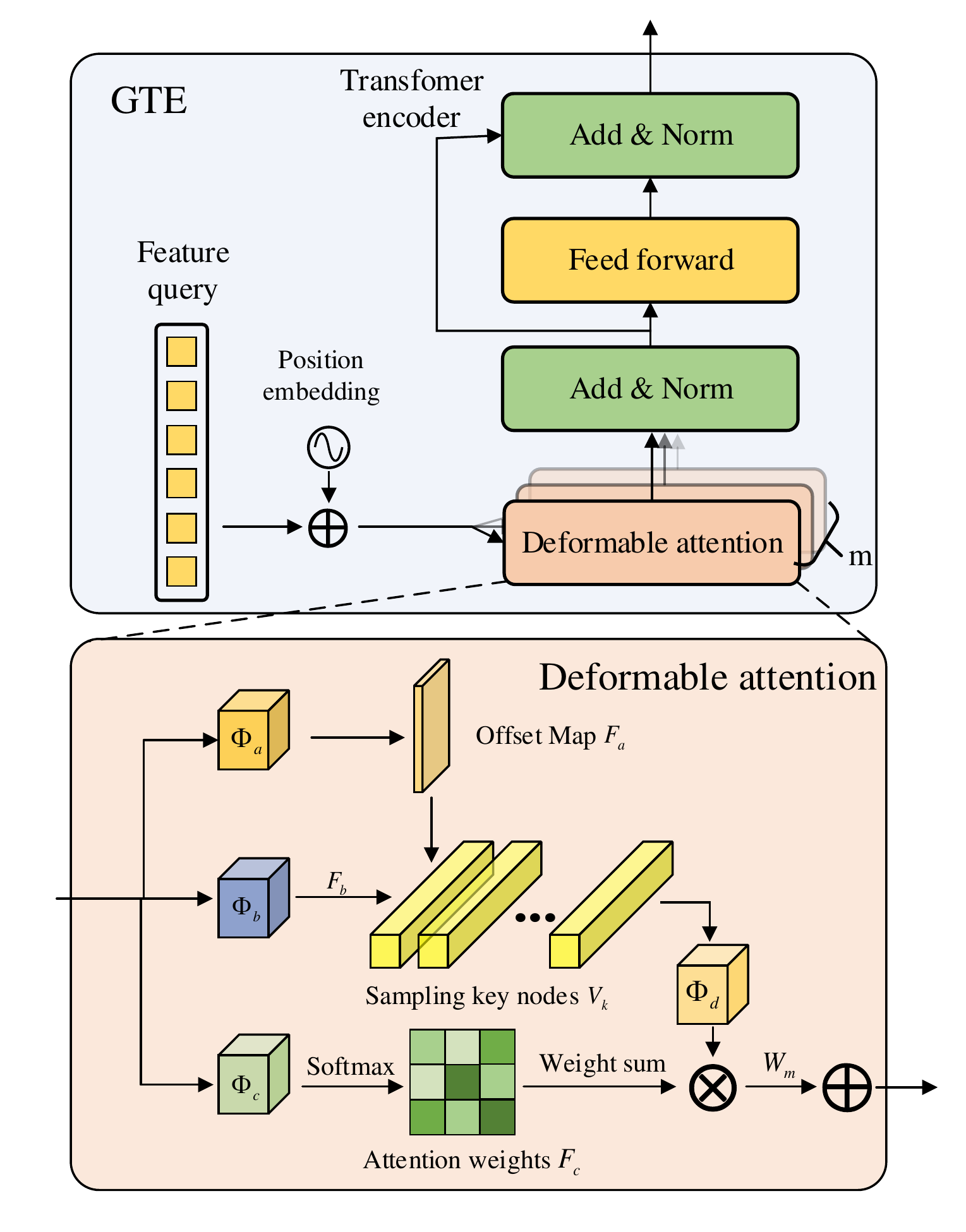}
    \caption{Diagram of the proposed GTE Module (FC: fully connected layer).}
    \vspace{-0.6cm}
\end{figure}

To bridge this gap, we attempt to utilize the power of global attention, which considers the interaction among all pixels. Nevertheless, global attention would result in serious computing burden, which limits the depth of networks and resolution of input feature maps.

With the target of addressing this restriction, we resort to deformable attention to capture the structural content. Unlike global attention, deformable attention can self-adaptively detect valuable key samples and avoid calculating the similarity between query nodes and all values in feature maps. This strategy successfully reduces the computing complexity from $O(H^{2}W^{2}C)$ to $O(HWC)$.

Furthermore, we propose the GTE module through combining the advantages of deformable attention and Transformer encoder as illustrated in Fig. 3. Integrated with the outstanding inference capability of Transformer and self-adaptive global receptive field of deformable attention, GTE produces representative embeddings.

In the following, we will introduce the details of two components of GTE, Transformer encoder and deformable attention, respectively.

\vspace{2mm}
\noindent \textbf{Transformer encoder.} Transformer \cite{vaswani2017attention} is first proposed for natural language processing and then extensively applied to various computer vision tasks \cite{meinhardt2021trackformer}. Standard Transformer mainly comprises two components, an encoder and a decoder. In GTE, we build a network with a structure similar to the Transformer encoder to obtain powerful embeddings for subsequent association operations.

As shown in Fig. 3, transformer encoder typically consists of a multi-head attention block and one feed-forward network (FFN). Generally, given a query $q$ and a set of key elements $\Omega_{k}$ as input, Transformer first produces the relation maps via dot-product between $q$ and $k$ ($k \in \Omega_{k}$). Then, the obtained relation maps are normalized and correlated with $k$ again to generate representative embeddings. Afterwards, FFN is utilized to further extract the information in the embeddings.

Mathematically, the aforementioned procedure could be formulated as
\begin{align}
& \Phi_{T}(q, k) = \Gamma (\sum\limits_{i=1}^{N_{head}}W_{i}( \sum\limits_{j \in \Omega_{k}}A_{ij}W_{i}^{'}k_{j} ))  \\
& A_{ij} \propto exp(\frac{q^{T}U^{T}_{i}V_{i}k_{j}}{\rho})
\end{align}
where $W_{i}$, $W^{'}_{i}$, $U_{i}$ and $V_{i}$ are learnable weights. $\Phi_{T}(\cdot)$, $\Gamma(\cdot)$, $N_{head}$ and $\rho$ represent the Transformer, FFN, number of attention heads and normalization factor, respectively.

According to the above description, the computational complexity of Transformer encoder is $O(H^{2}W^{2}C)$ for input data with the shape of $(H, W, C)$. Therefore, the computing cost grows quadratically with respect to the expansion of image size, and the cost is mainly caused by the dot-product attention of multi-head attention. In this work, we adopt a novel attention strategy to alleviate the enormous calculation burden.

\vspace{2mm}
\noindent \textbf{Deformable attention.} As mentioned before, the huge computational cost of global attention results in slow convergence and limited image resolution during the training phase. In order to overcome this problem, we employ deformable attention instead of global attention.

\begin{figure}[tbp]
    \vspace{-0.2cm}
    \centering
    \includegraphics[scale=0.5]{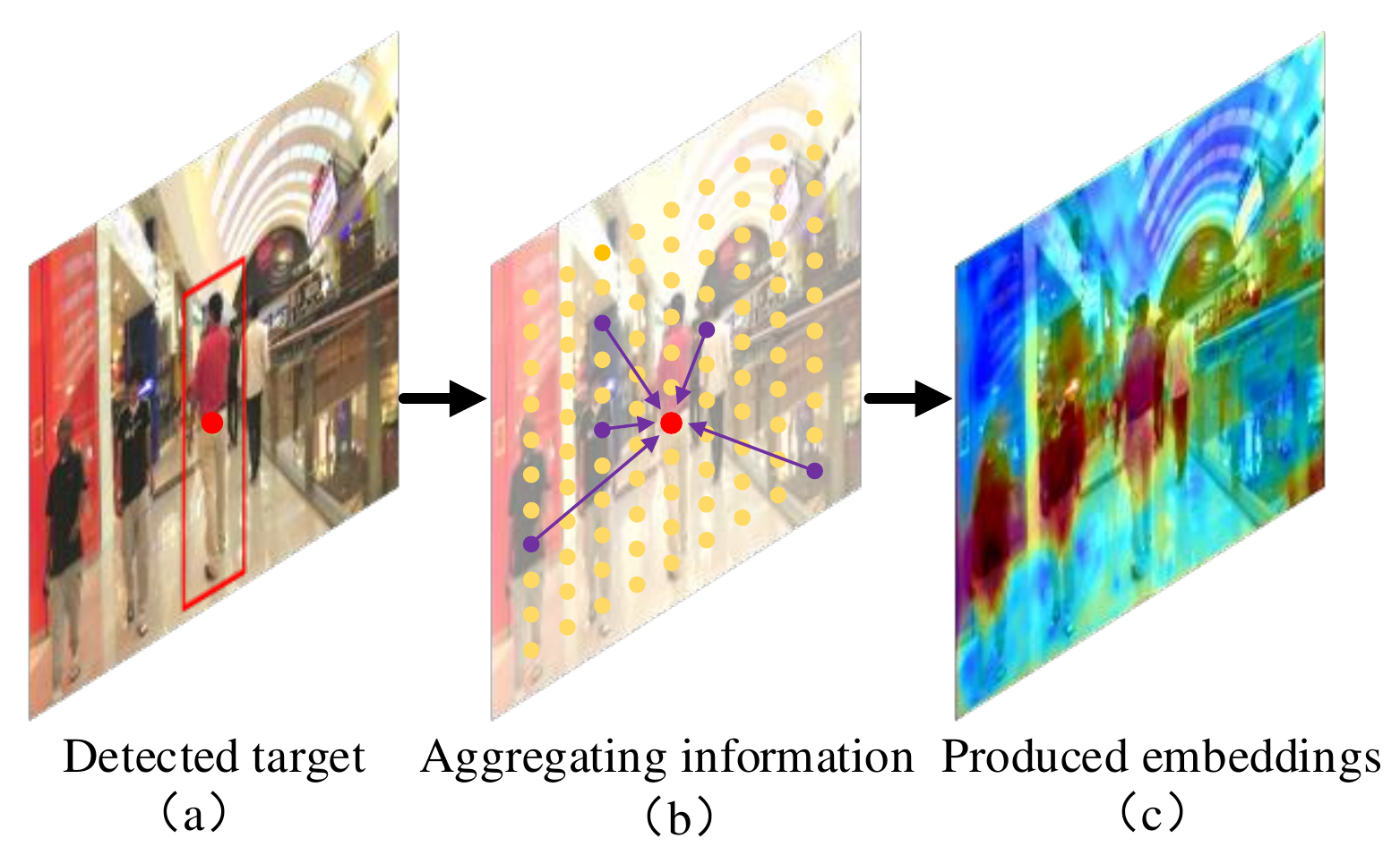}
    \caption{Abstract diagram of deformable attention. The red point is the query node and purple points are the self-adaptively selected key samples. In contrast to global encoder, deformable attention avoids the huge computational burden of considering the relation among all pixels.}
    \vspace{-0.6cm}
\end{figure}

Fig. 4 presents the basic idea behind deformable attention. For any query node in the detected regions of interest (Fig 4(a)), deformable attention self-adaptively selects valuable key samples across the whole image (Fig 4(b)). Afterwards, discriminative representation is produced (Fig 4(c)) through interacting information between query nodes and the corresponding key samples.

The details of deformable attention are illustrated in Fig. 3. First of all, given input feature maps $I$, three independent encoders, $\Phi_{a}(\cdot)$, $\Phi_{b}(\cdot)$ and $\Phi_{c}(\cdot)$, severally encode the input as the offset maps $F_{a}$, key maps $F_{b}$ and query attention maps $F_{c}$. Notably, if we select $N_{k}$ key samples for every query node, $F_{a}$ would contain $2N_{k}$ channels, which are horizontal and vertical coordinate offsets of the $N_{k}$ key samples relative to the corresponding query node. Hence, for every query node $q \in I$, we can know its coordinate $Z_{q}$ and the offsets of key samples $\triangle Z_{k}=\{\triangle Z_{k}^{i}\}_{i=1}^{N_{k}}$ relative to $Z_{q}$ based on $F_{a}$. Then, the coordinates of key samples $Z_{k}=\{Z_{k}^{i}\}_{i=1}^{N_{k}}$ can be computed as follows.
\begin{align}
Z^{i}_{k} = Z_{q} + \triangle Z_{k}^{i}
\end{align}

Afterwards, according to the coordinates of selected key samples $Z_{k}=\{Z^{i}_{k}\}_{i=1}^{N_{k}}$ and key maps $F_{b}$, we obtain the key sample vectors $V_{k} = \{V_{k}^{i}\}_{i=1}^{N_{k}}$. They are further transformed as $\tilde{V}_{k}$ by the encoder $\Phi_{d}(\cdot)$. Moreover, we crop the query attention vectors $V_{q}=\{V_{q}^{i}\}_{i=1}^{N_{k}}$ from $F_{c}$ with respect to $Z_{k}$. The final output maps $F_{o}$ can be calculated as:

\begin{align}
F_{o} = W_{m} \sum\limits_{i=1}^{N_{k}} V_{q}^{i} \bullet F_{c}^{i}
\end{align}
where $W_{m}$ denotes trainable parameters and $\bullet$ is the Hadamard multiplication \cite{muller1992hadamard}. According to the described procedure, the computational complexity of deformable attention is $O(HWC)$ compared with global attention, the complexity of which is $O(H^{2}W^{2}C)$.

\vspace{-2mm}
\subsection{Detection and Association}
We devise a detection module $\Psi_{d}(\cdot)$ similar to Centernet to localize the targets. Given the decoupled detection-specific representation, $\Psi_{d}(\cdot)$ would detect the center points of concerned objects, regress the corresponding center offsets and estimate the bounding box shapes simultaneously. Combing all the obtained outputs, the regions that contain targets are determined.

Afterwards, based on the embeddings produced by GTE and estimated bounding boxes, we employ the Hungarian algorithm to match objects among various frames and generate the trajectories. We keep the same settings in FairMOT \cite{zhan2020simple}. First of all, we initialize tracklets based on the detected boxes in the first frame. Then, in the subsequent frames, we link the detected boxes to the original tracklets according to the cosine distances between the corresponding computed embedding vectors. If there exist unmatched detected boxes, they will be connected to newly initialized trajectories. In addition, we adopt the trajectory filling strategy mentioned in MAT \cite{han2020mat} to balance the false positive and false negative scores.

\vspace{-2mm}
\subsection{Optimization objectives}
Since RelationTrack comprises several subtasks, we leverage multiple optimization objectives to train its various parts. These optimization objectives are introduced as follows.

\vspace{2mm}
\noindent \textbf{Detection branch.} To the end of localizing the concerned objects, the detection branch first estimates the center points of targets. Denote the $i_{\rm th}$ bounding box annotation in a frame as $b^{i}$, and its corresponding upper left and lower right coordinates are $(l^{i}, t^{i})$ and $(r^{i}, b^{i})$, respectively. The center point of this bounding box could be expressed as $(c_{x}^{i}, c_{y}^{i})$, where $c_{x}^{i} = \frac{l^{i} + r^{i}}{2}$, $c_{y}^{i} = \frac{t^{i} + b^{i}}{2}$. Assuming there are totally $N$ bounding box annotations in this frame, we can produce the heatmap groundtruth $\hat{R}$ as follows
\begin{align}
\hat{R}_{xy} = \sum\limits_{i=1}^{N} exp(-\frac{(x-c_{x}^{i})^{2} + (y - p_{y}^{i})^{2}}{2(\sigma_{p})^{2}})
\end{align}
where $\hat{R}_{xy}$ is the heatmap pixel with the coordinate of $(x, y)$ and $\sigma_{p}$ represents the standard deviation value that is self-adaptively adjusted according to the heatmap scale. Denoting the estimated heatmap pixel at $(x, y)$ as $R_{xy}$, the similarity value for regressing this pixel can be defined as
\begin{align}
  L^{h}_{xy} = \begin{cases}
  (1 - R_{xy})^{\alpha} logR_{xy}, \hat{R}_{xy}=1  \\
  (1 - \hat{R}_{xy})^{\beta}(R_{xy})^{\alpha}log(1 - R_{xy}), \hat{R}_{xy} \neq 1
  \end{cases}
\end{align}
where $\alpha$ and $\beta$ are hyper-parameters \cite{lin2017focal}. Afterwards, the loss function for estimating the center points of targets is formulated as follows.
\begin{align}
L^{h} = -\frac{1}{N} \sum\limits_{y=1}^{H}\sum\limits_{x=1}^{W} L^{h}_{xy}
\end{align}

In order to determine the bounding box regions, we build another network branch to estimate the box shapes and offsets. In our implementation, the label of the $i_{\rm th}$ box shape is expressed as $\hat{s}^{i} = (r^{i} - l^{i}, b^{i} - t^{i})$ and its corresponding offset label is $\hat{o}^{i} = (\frac{c_{x}^{i}}{4} - \lfloor\frac{c_{x}^{i}}{4}\rfloor, \frac{c_{y}^{i}}{4} - \lfloor\frac{c_{y}^{i}}{4}\rfloor)$, where $\lfloor \cdot \rfloor$ represents an operator that rounds down the input decimals. The optimization objective of this branch for predicting bounding boxes can be given as
\begin{align}
L^{b} = \sum\limits_{i=1}^{N} \Vert o^{i} - \hat{o}^{i} \Vert_{1} + \vert s^{i} - \hat{s}^{i} \Vert_{1}
\end{align}
where $o^{i}$ and $s^{i}$ are the output of networks, and $\Vert \cdot \Vert_{1}$ denotes the $l_{1}$ measurement \cite{li2020deep}.

\vspace{2mm}
\noindent \textbf{ReID branch.} We regard the ReID task as a classification problem and the targets with an identical identity belong to the same category. Given multiple bounding boxes, the networks of the ReID branch would transform the features in every bounding box to a class distribution vector $p=\{p_{i}\}^{K}_{i=1}$, where $K$ denotes the total number of categories. Assuming the one-hot annotation for the ReID task is $q=\{q_{j}\}^{K}_{j=1}$, the loss function adopted by the ReID branch could be formulated as follows.
\begin{align}
L^{r} = - \sum\limits_{j=1}^{K} \sum\limits_{i=1}^{K} q_{j}log(p_{i})
\end{align}

\vspace{2mm}
\noindent \textbf{Overall optimization objective.} Combining the above loss functions with learnable coefficients $\omega_{1}$ and $\omega_{2}$, we can obtain the overall optimization objective $L$ for RelationTrack, which is given as follows.
\begin{align}
& L^{d} = L^{h} + L^{b} \\
L = \frac{1}{2}(\frac{1}{e^{\omega_{1}}} & L^{d} + \frac{1}{e^{\omega_{2}}}L^{r} + \omega_{1} + \omega_{2})
\end{align}

\section{Experiment}
This section presents the experimental details. Specifically, Subsection A introduces the adopted training and evaluation datasets as well as the evaluation metrics. Among the datasets, MOT15, MOT16 and MOT17 are used for validating the models. Subsection B presents the implementation details in the experiments. Afterwards, Subsection C demonstrates the superiority of RelationTrack by comparing it with existing state-of-the-art MOT methods. Subsection D proves the effectiveness of various components in RelationTrack through ablation experiments. Moreover, Subsection E and F indicate that the proposed GCD and GTE modules can enhance the tracking precision efficiently. Finally, Subsection G reveals the robustness of RelationTrack towards extreme cases.

\vspace{-2mm}
\subsection{Datasets and evaluation metrics}

\noindent \textbf{MOT15.} MOT15 \cite{leal2015motchallenge} is the first release of MOTChallenge and it comprises 22 sequences, a half for training and the other half for testing. This dataset totally contains 996 seconds of videos, which include 11286 frames.

\vspace{2mm}
\noindent \textbf{MOT16.} MOT16 \cite{milan2016mot16} is a commonly adopted benchmark in MOT. Composed of 14 sequences, it covers various scenarios, viewpoints, camera poses and weather conditions. Similar to MOT15, 7 sequences in MOT16 are for training and the others are for validation.

\vspace{2mm}
\noindent \textbf{MOT17.} MOT17 \cite{milan2016mot16} is established through reconstructing MOT16. In contrast to MOT16, MOT17 provides more reliable groundtruth and more detection bounding boxes produced by various detectors, which include DPM \cite{felzenszwalb2009object}, SDP \cite{yang2016exploit}, Faster RCNN \cite{ren2016faster}. The rest is the same as MOT16.

\vspace{2mm}
\noindent \textbf{MOT20.} Compared with aforementioned datasets, MOT20 \cite{dendorfer2020mot20} is more challenging. It consists of 8 video sequences captured in 3 very crowded scenes. In some frames, more than 220 pedestrians are included. Meanwhile, the data in MOT20 presents high diversity, which could be indoor or outdoor, at day or night.

\vspace{2mm}
\noindent \textbf{Extended datasets.} Following the settings of preceding works \cite{zhan2020simple}, besides the above benchmarks on MOT Challenge, we adopt some other datasets for training, which include ETH \cite{ess2008mobile}, CityPerson \cite{zhang2017citypersons}, CalTech \cite{dollar2009pedestrian}, CUHK-SYSU \cite{xiao2017joint}, PRW \cite{zhong2018camstyle} and CrowdHuman \cite{shao2018crowdhuman}. Meanwhile, the performance verification and analysis experiments are mainly performed on MOT16, MOT17 and MOT20.

\vspace{2mm}
\noindent \textbf{Evaluation metrics.} The verification of RelationTrack is carried out based on the CLEAR-MOT Metrics \cite{bernardin2008evaluating}, which is a commonly adopted metric set. It is composed of ID F1 score (IDF1), higher order tracking accuracy (HOTA), multiple object tracking accuracy (MOTA), multiple object tracking precision (MOTP), mostly tracked rate (MT), mostly lost rate (ML), false positives (FP), false negatives (FN), identity switches (IDS) and inference speed (IS). Among them, HOTA \cite{luiten2021hota} is a recently proposed metric. In contrast to former metrics, it reflects the capability of tracking, detection and association simultaneously. \textbf{IDF1, HOTA, MOTA are the most primary indexes for indicating the performance of evaluated models}.

\begin{table*}[ht]
    \vspace{-0.2cm}
    \centering
    \caption{Comparison with preceding state-of-the-art methods on MOT16.}
    \resizebox{150mm}{17mm}{
    \begin{tabular}{ccccccccccc}
    \hline \hline
    Model & IDF1$\uparrow$ & HOTA$\uparrow$ & MOTA$\uparrow$ & MOTP$\uparrow$ & MT$\uparrow$ & ML$\downarrow$ & FP$\downarrow$ & FN$\downarrow$ & IDS$\downarrow$ & IS$\uparrow$ \\
    \cline{1-11}
    JDE \cite{wang2019towards} & 55.8 & - & 64.4 & - & 35.4\% & 20.0\% & - & - & 1544 & 18.8 \\
    CTracker \cite{peng2020chained} & 57.2 & 48.8 & 67.6 & 78.4 & 32.9\% & 23.1\% & \textbf{8934} & 48305 & 1117 & \textbf{34.4} \\
    TubeTK \cite{pang2020tubetk} & 59.4 & 48.7 & 64.0 & 78.3 & 33.5\% & 19.4\% & 10962 & 53626 & 4137 & 1.0 \\
    DeepSortv2 \cite{wojke2017simple} & 62.2 & 50.1 & 61.4 & 79.1 & 32.8\% & 18.2\% & 12852 & 56668 & 2008 & 17.4 \\
    MAT \cite{han2020mat} & 63.8 & 54.4 & 70.5 & 80.4 & \textbf{44.7\%} & 17.3\% & 11318 & 41592 & 928 & 9.1  \\
    CSTrack \cite{liang2020rethinking} & 71.8 & - & 70.7 & - & 38.2\% & 17.8\% & - & - & 1071 & 15.8 \\
    FairMOTv2 \cite{zhan2020simple} & 72.8 & 59.8 & 74.9 & \textbf{81.2} & 44.7\% & \textbf{15.9\%} & 10163 & 34484 & 1074 & 25.4 \\
    $\textbf{RelationTrack (ours)}$ & \textbf{75.8} & \textbf{61.7} & \textbf{75.6} & 80.9 & 43.1\% & 21.5\% & 9786 & \textbf{34214} & \textbf{448} & 7.4 \\
    \hline \hline
    \end{tabular}}
    \vspace{-0.3cm}
\end{table*}

\begin{table*}[ht]
    \vspace{-0.2cm}
    \centering
    \caption{Comparison with preceding state-of-the-art methods on MOT17.}
    \resizebox{150mm}{16mm}{
    \begin{tabular}{ccccccccccc}
    \hline \hline
    Model & IDF1$\uparrow$ & HOTA$\uparrow$ & MOTA$\uparrow$ & MOTP$\uparrow$ & MT$\uparrow$ & ML$\downarrow$ & FP$\downarrow$ & FN$\downarrow$ & IDS$\downarrow$ & IS$\uparrow$ \\
    \cline{1-11}
    CTracker \cite{peng2020chained} & 57.4 & 49.0 & 66.6 & 78.2 & 32.2\% & 24.2\% & 22284 & 160491 & 5529 & 6.8 \\
    TubeTK \cite{pang2020tubetk} & 58.6 & 48.0 &63.0 & 78.3 & 31.2\% &  19.9\% & 27060 &  177483 & 5529 & 6.8 \\
    MAT \cite{han2020mat} & 63.1 & 53.8 & 69.5 & 80.4 & \textbf{43.8\%} & 18.9\& & 30660 & 138741 & 2844 & 9.0 \\
    CenterTrack \cite{zhou2020tracking} & 64.7 & 52.2 & 67.8 & 78.4 & 34.6\% & 24.6\% & \textbf{18489} & 160332 & 3039 & 22.0 \\
    CSTrack \cite{liang2020rethinking} & 71.6 & - & 70.6 & - & 37.5\% & 18.7\% & - & - & 3465 & 15.8 \\
    FairMOTv2 \cite{zhan2020simple} & 72.3 & 59.3 & 73.7 & \textbf{81.3} & 43.2\% & \textbf{17.3\%} & 27507 & \textbf{117477} & 3303 & \textbf{25.9} \\
    $\textbf{RelationTrack (ours)}$ & \textbf{74.7} & \textbf{61.0} & \textbf{73.8} & 81.0 & 41.7\% & 23.2\% & 27999 & 118623 & \textbf{1374} & 7.4 \\
    \hline \hline
    \end{tabular}}
    \vspace{-0.3cm}
\end{table*}

\begin{table*}[ht]
    \vspace{-0.2cm}
    \centering
    \caption{Comparison with preceding state-of-the-art methods on MOT20.}
    \resizebox{150mm}{11mm}{
    \begin{tabular}{ccccccccccc}
    \hline \hline
    Model & IDF1$\uparrow$ & HOTA$\uparrow$ & MOTA$\uparrow$ & MOTP$\uparrow$ & MT$\uparrow$ & ML$\downarrow$ & FP$\downarrow$ & FN$\downarrow$ & IDS$\downarrow$ & IS$\uparrow$ \\
    \cline{1-11}
    MLT \cite{zhang2020multiplex} & 54.6 & 43.2 & 48.9 & 78.0 & 30.9\% & 22.1\% & 45660 & 216803 & \textbf{2187} & 3.7 \\
    FairMOTv2 \cite{zhan2020simple} & 67.3 & 54.6 & 61.8 & 78.6 & 68.8\% & 7.6\% & 103440 & \textbf{88901} & 5243 & \textbf{13.2} \\
    CSTrack \cite{liang2020rethinking} & 68.6 & 54.0 & 66.6 & 78.8 & 50.4\% & 15.5\% & \textbf{25404} & 144358 & 3196 & 4.5 \\
    $\textbf{RelationTrack (ours)}$ & $\textbf{70.5}$ & $\textbf{56.5}$ & \textbf{67.2} & \textbf{79.2} & \textbf{62.2\%} & \textbf{8.9\%} & 61134 & 104597 & 4243 & 2.7 \\
    \hline \hline
    \end{tabular}}
    \vspace{-0.3cm}
\end{table*}

\vspace{-3mm}
\subsection{Implementation Details}
In our experiments, we employ DLA-34 pre-trained on the COCO dataset \cite{lin2014microsoft} and fine-tuned on the aforementioned datasets as the backbone of RelationTrack. Its parameters are updated using the Adam optimizer \cite{kingma2015adam} with the initial learning rate of $10^{-4}$. During the training procedure, the input batch size is set as 12 and the resolution of every image is 1088$\times$608. The experiments are conducted on 2 NVIDIA GeForce RTX 2080Ti GPUs.

\vspace{-4mm}
\subsection{Comparison with preceding SOTAs}
In this part, we compare the performance of RelationTrack with preceding SOTA methods on three widely adopted benchmarks, i.e., MOT16, MOT17 and MOT20. The results are reported in Table I, Table II and Table III, respectively. As shown in these three tables, RelationTrack has come out among the top in various metrics and surpassed the contrasted counterparts by large margins, especially on the IDF1, HOTA, MOTA and IDS metrics.

\textbf{MOT16/17.} For instance, according to Table I and Table II, RelationTrack obtains the metric IDF1 of 75.8\% on MOT16 and 74.7\% on MOT17. It outperforms the recently proposed FairMOTv2 \cite{zhan2020simple} by 3.0\% ($75.8\% - 72.8\% $) and 2.4\% ($74.7\% - 72.3\%$) on MOT16 and MOT17, respectively. Meanwhile, RelationTrack also behaves better on MOTA than most other trackers. The results indicate the outstanding tracking capability of RelationTrack, which is because that the GCD and GTE modules can produce discriminative features while maintaining competitive detection accuracies. Meanwhile, it can be observed from Table I and Table II that RelationTrack also behaves well on the metric IDS. This phenomenon reveals that the tracking trajectories produced by RelationTrack are still stable even in large-scale and complex scenes.

\textbf{MOT20.} To further evaluate the proposed framework, we verify its performance on the MOT20 benchmark. As shown in Table III, RelationTrack behaves the best on most metrics. Particularly, it surpasses FairMOTv2 by 3.2\% ($70.5\% - 67.3\%$) on IDF1, 1.9\% ($56.5\% - 54.6\%$) on HOTA and 5.4\% ($67.2\% - 61.8\%$) on MOTA.

\vspace{-4mm}
\subsection{Ablation Study}

In this part, we analyze the effectiveness of GCD and GTE through ablation experiments on MOT17. To this end, we adopt the RelationTrack without GCD and GTE as the baseline model. The results are presented in Table IV.

According to the $2_{\rm nd}$ and $3_{\rm rd}$ rows of Table IV, the proposed GCD module enhances the tracking performance of the baseline by 1.6\% ($74.9\%-73.3\%$) on IDF1 and 0.9\% ($69.4\% - 68.5\%$) on MOTA. These improvements confirm the importance of alleviating the aforementioned optimization contradiction by decoupling features.

\begin{table}[htbp]
    \vspace{-0.2cm}
    \caption{The effectiveness of proposed blocks.}
    \centering
    \resizebox{85mm}{11mm}{
    \begin{tabular}{cccccc}
    \hline \hline
    Model & IDF1$\uparrow$ & MOTA$\uparrow$ & FP$\downarrow$ & FN$\downarrow$ & IDS$\downarrow$ \\
    \cline{1-6}
    Baseline & 73.3 & 68.5 & 2567 & 14135 & 293 \\
    \cline{1-6}
    Baseline + GCD & 74.9 & 69.4 & 2791 & 13477 & \textbf{286} \\
    Baseline + GTE & 73.6 & 69.5 & \textbf{2242} & 14013 & 326 \\
    RelationTrack & \textbf{75.3} & \textbf{70.1} & 2583 & \textbf{13301} & 309 \\
    \hline \hline
    \end{tabular}}
    \vspace{-0.2cm}
\end{table}

\begin{figure*}[ht]
    \centering
    \includegraphics[scale=0.24]{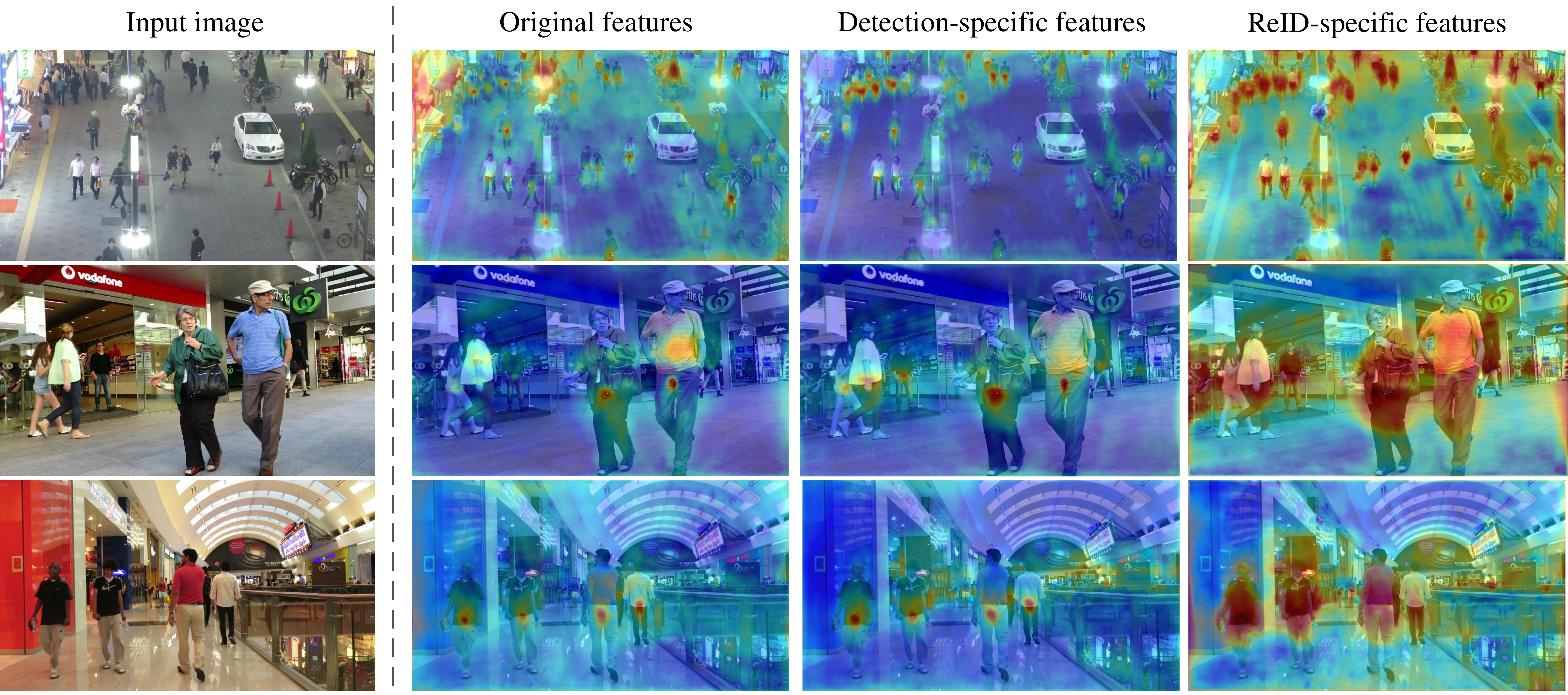}
    \caption{Visualization of the original and decoupled representation (the detection-specific and ReID-specific features).}
    \vspace{-0.2cm}
\end{figure*}

Meanwhile, it could be observed that GTE also benefits the inference procedure. As shown, the baseline with GTE outperforms the pure baseline by 0.3\% ($73.6\% - 73.3\%$) on IDF1 and 1.0\% ($69.5\% - 68.5\%$) on MOTA.

Incorporating the baseline model with both GCD and GTE, the resulted RelationTrack achieves outstanding tracking precision. As presented in Table IV, RelationTrack surpasses the baseline by 2.0\% ($75.3\% - 73.3\%$) on IDF1 and 1.6\% ($70.1\% - 68.5\%$) on MOTA. The results indicate the great power brought through combining GCD and GTE.

\vspace{-2mm}
\subsection{Visualization of GCD}

This part aims to verify whether GCD really addresses the optimization contradiction between detection and ReID through decoupling features. To realize this target, we visualize and compare the original and decoupled representation, which is illustrated as Fig. 5.

As shown, in the original feature maps, the model ignores many small yet important targets. Besides, some irrelevant areas are concentrated on mistakenly. On the contrary, when the features are decoupled, the center parts of targets are highlighted in the detection-specific feature maps. Meanwhile, only the regions that cover pedestrians are focused on in the ReID-specific embeddings. This phenomenon demonstrates that GCD has decoupled the representation vectors as designed. Correspondingly, the optimization contradiction between the detection and ReID branches is addressed successfully.

\vspace{-2mm}
\subsection{Impact of key sample numbers}

As introduced in Section V, instead of utilizing global attention, we employ deformable attention to capture the long-range dependency. Rather than analyzing the relationship between query nodes and all the other pixels, deformable attention only considers a small amount of adaptively selected key samples. Therefore, the number of the key samples could influence the final tracking precision of models. In this part, we aim to study this problem and the corresponding results are reported in Table V.

As shown in the $4_{\rm th}$ column of Table V, when more than 9 samples are produced, the results measured by the metric MOTA do not vary obviously. On the contrary, presented in the $3_{\rm rd}$ column, the performance of networks under the metric IDF1 is affected significantly. When the number of key samples are 6, 9, 12 and 15, the corresponding IDF1 scores are 73.9\%, 75.3\%, 75.0\% and 74.9\%, respectively. Although we can further incorporate more key samples, it would result in more computing resource demand. Considering both the tracking performance and calculation burden, we decide to select 9 key samples for every query node.

\begin{table}[htbp]
    \vspace{-0.2cm}
    \caption{Comparison of different number of sampling key nodes.}
    \centering
    \resizebox{78mm}{12mm}{
    \begin{tabular}{c|c|ccccc}
    \hline \hline
    \multirow{2}{*}{Module} & \multirow{2}{*}{Num} & \multicolumn{5}{c}{MOT17} \\
    \cline{3-7}
    & & IDF1$\uparrow$ & MOTA$\uparrow$ & FP$\downarrow$ & FN$\downarrow$ & IDS$\downarrow$ \\
    \cline{1-7}
    \multirow{3}{*}{GTE} & 6 & 73.9 & 69.6 & 2576 & 13485 & 378 \\
    & 9 & \textbf{75.3} & \textbf{70.1} & 2583 & 13301 & 309\\
    & 12 & 75.0 & 70.0 & 2636 & \textbf{13261} & \textbf{290}\\
    & 11 & 74.9 & 70.0 & \textbf{2262} & 13652 & 292\\
    \hline \hline
    \end{tabular}}
    \vspace{-0.3cm}
\end{table}

\begin{figure}[htbp]
    \centering
    \includegraphics[scale=0.3]{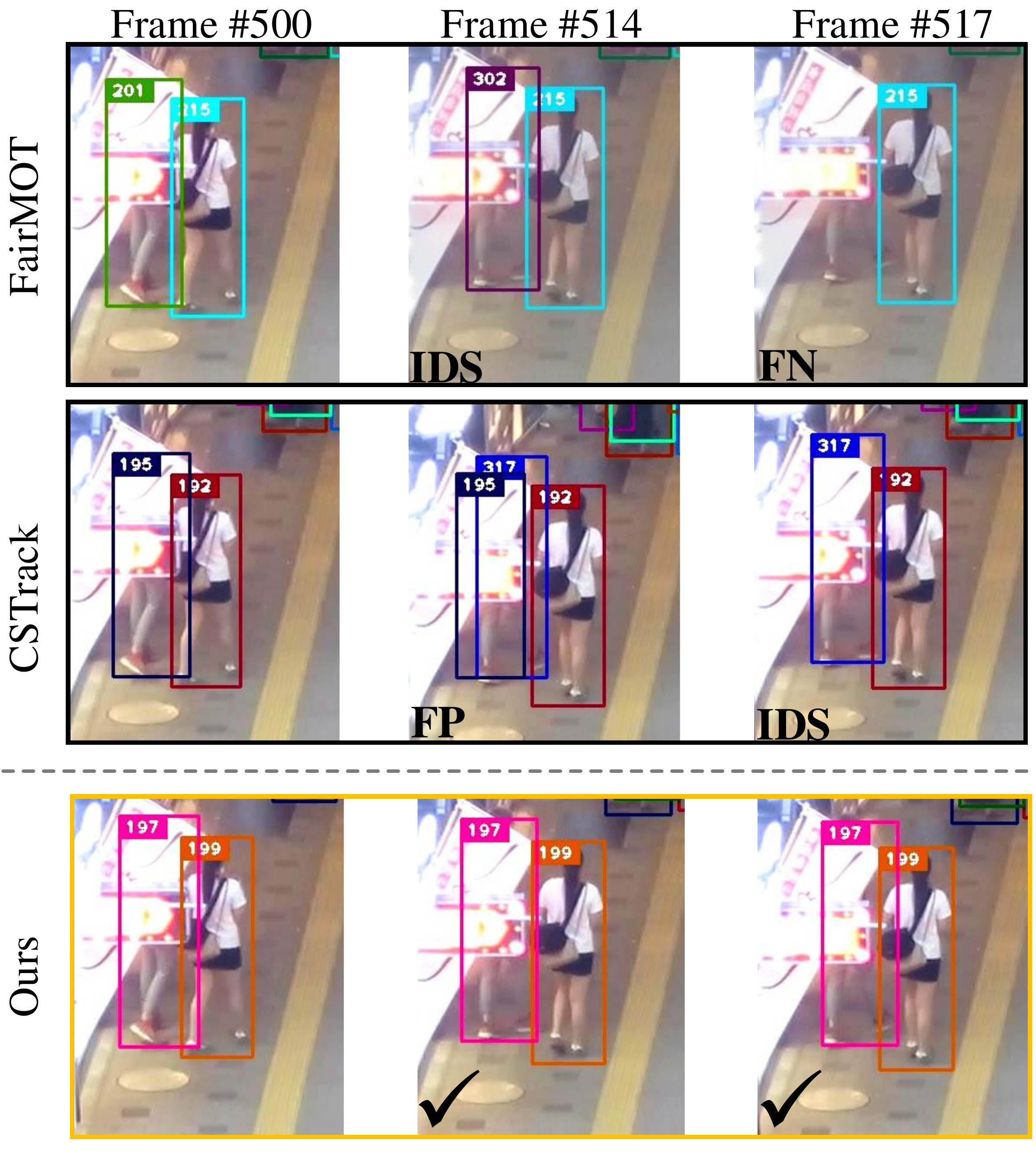}
    \caption{Robustness analysis of RelationTrack compared with the FariMOT and CSTrack frameworks. The check mark indicates that the results are correct. IDS, FP and FN denote different kinds of false estimations which include identity switch, false negative and false positive, respectively.}
    \vspace{-0.4cm}
\end{figure}

\begin{figure*}[htbp]
    \centering
    \includegraphics[scale=0.5]{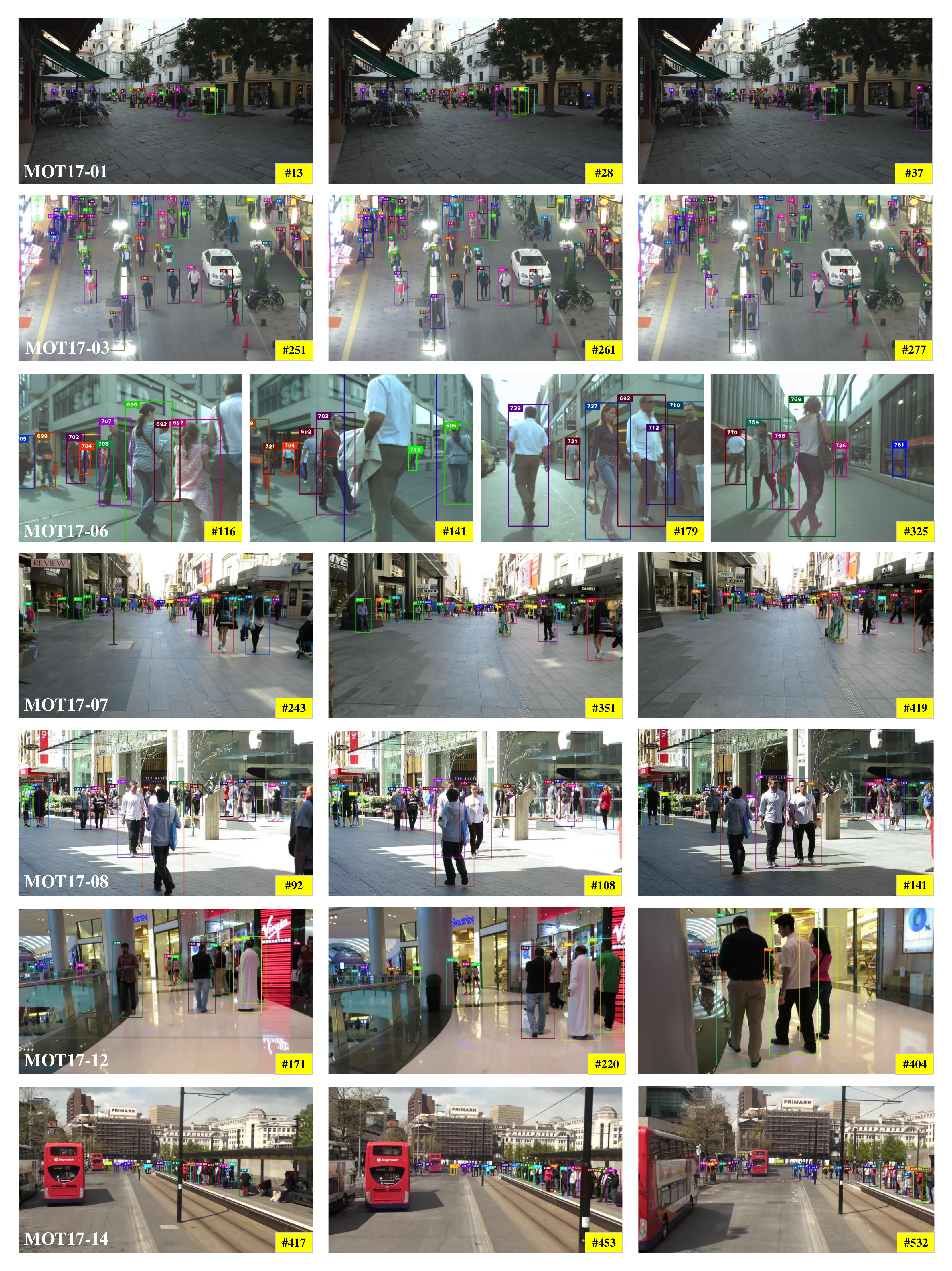}
    \caption{Tracking examples of RelationTrack on the MOT17 dataset.}
    \vspace{-0.4cm}
\end{figure*}

\vspace{-2mm}
\subsection{Robustness analysis}

In this part, we analyze the robustness of RelationTrack under extreme cases. A hard case is given in Fig. 6 as an example. In this figure, a person is partly occluded and hard to be detected. Both FairMOT and CSTrack fail to extract representative embeddings and associate it with the correct trajectory. Specifically, FairMOT labels this person with a false identity in Frame \#514 and overlooks this region in Frame \#517. Likewise, CSTrack produces wrong results in the two frames. On the contrary, RelationTrack identifies the target successfully. This example reveals the robustness of RelationTrack due to its strong capability of extracting features.

In order to further confirm this issue, we illustrate more cases in Fig. 7. These cases have covered various practical tracking situations, which include indoor and outdoor scenarios, day and night periods, huge and small targets, etc. The outstanding estimated results prove that RelationTrack can achieve robust and precise performance even under those difficult conditions.

\section{Conclusion}
In this work, we observed that the optimization contradiction between the detection and ReID branches had restricted current MOT methods from further improvements. Correspondingly, we developed a module named GCD that alleviates this contradiction through decoupling the features as detection-specific and ReID-specific ones. Moreover, we noticed that preceding MOT frameworks mostly only utilize the local features and neglect to consider the global semantic relation among pixels. We attempted to bridge this gap by devising a network similar to the Transformer encoder. Nevertheless, this strategy suffers from heavy computation burden. To address this issue, we replaced the global attention operator in Transformer encoder as deformable attention and designed a novel module named GTE. This module can capture the global structural information while only consuming a limited amount of calculation resources. Combining the GCD and GTE modules, we proposed a competitive MOT framework, namely RelationTrack. Its performance has been validated through 5 groups of experiments on 3 classic MOT benchmark datasets, which include MOT16, MOT17 and MOT20. The results indicate that RelationTrack has outperformed the contrasted counterparts significantly and established new SOTA results.

\ifCLASSOPTIONcaptionsoff
  \newpage
\fi

\bibliographystyle{IEEEtran}
\bibliography{references}
\end{document}